\documentclass[sigconf, natbib=false, language=english, urlbreakonhyphens=false]{acmart}
\usepackage[datamodel=acmdatamodel, style=acmnumeric, backend=biber]{biblatex}
\usepackage{multirow}
\usepackage[ruled, linesnumbered]{algorithm2e}
\usepackage{mdframed}
\usepackage{fancyvrb}

\AtBeginDocument{%
  }

\makeatletter
\def\acm@mailurl#1{\href{mailto:#1}{#1}}
\patchcmd{\@mkauthors@iii}{\nolinkurl}{\acm@mailurl}{}{}
\patchcmd{\@mkauthors@iii}{\nolinkurl}{\acm@mailurl}{}{}
\makeatother

\setcopyright{rightsretained}
\copyrightyear{2026}
\acmDOI{10.70314/is.2026.aihc.42}

\acmConference[Information Society 2026]{Information Society 2026: 29th international multiconference}{5--9 October 2026}{Ljubljana, Slovenia}
\acmISBN{}
\begin{document}

\title[Detecting Spin in Clinical Trials with LLMs]{Detecting Spin in Clinical Trials with Large Language Models}

\author{Tjaš Ajdovec}
\affiliation{%
  \institution{University of Ljubljana, Faculty of Computer and Information Science}
  %\city{Ljubljana}
  \country{Slovenia}}
\email{tjas.ajdovec@gmail.com}

\author{Marko Robnik-Šikonja}
\affiliation{%
  \institution{University of Ljubljana, Faculty of Computer and Information Science}
  %\city{Ljubljana}
  \country{Slovenia}}
\email{marko.robnik@fri.uni-lj.si}

\author{Simon Šuster}
\affiliation{%
  \institution{Independent researcher}
  \city{Ljubljana}
  \country{Slovenia}}
\email{sim.suster@gmail.com}

\renewcommand{\shortauthors}{Ajdovec et al.}

\begin{abstract}

  Spin in clinical trials includes reporting practices that distort the presentation of results. This is particularly critical in medicine, where spin is present in more than 50\% of randomized controlled trials that fail to reach statistical significance. The comparison of primary and reported outcomes is crucial for detecting several types of spin, including outcome switching. We used 300 pairs of outcomes labeled with semantic similarity to develop a system for automatic detection of outcome switching. We evaluated baseline text similarity models and open-source LLMs using generated similarity scores and the Youden index to determine the classification threshold. The proposed approach involves prompt engineering, classification based on token probabilities, and majority voting for the final decision. The results on the test set of 2,496 examples with an $F_1$ score of 0.78 and an accuracy of 0.90 outperform baseline text similarity models but trail behind fine-tuned versions of BERT. We used LLMs to generate natural language explanations for the classified instances and manually assessed their quality.

\end{abstract}


\maketitle

%% BEGIN
\section{Introduction}

The phenomenon of \textbf{spin} covers several reporting practices that present study results in a distorted way, usually to the benefit of the authors. It is a long-known problem, especially in randomized controlled trials (RCTs), where reporting rules are clearly defined, yet incorrect reporting still occurs. This leads to redundant or misdirected follow-up research~\cite{glasziou2014reducing}. Several tools exist to help authors and readers automatically detect this kind of incorrect reporting.

Large language models (LLMs) have recently become capable of solving many natural language processing (NLP) tasks \textbf{without task-specific training}~\cite{naveed2025comprehensive}. Existing spin detection tools rely on dedicated models trained on expensive, manually labeled data, and to our knowledge none of them make use of LLMs. We investigate whether LLMs can be used to classify pairs of primary and reported clinical trial outcomes as similar or different. This is a core subtask needed for detecting a type of spin called \textbf{outcome switching} - compromised reporting of primary and secondary outcomes in the results.

We addressed the following \textbf{research questions} in the context of detecting outcome switching in RCT reports:
\begin{enumerate}
  \item How successful are LLMs, without additional training, at classifying pairs of clinical trial outcomes as similar or different?
  \item How do the results and behavior of different LLMs differ with respect to generality and comprehension ability?
  \item Can LLMs generate meaningful, human-readable explanations for the already classified examples?
\end{enumerate}
We report the work in six sections. Section 2 outlines the related work. Section 3 presents the methodology and Section 4 the results. The discussion is contained in Section 5, and the conclusions follow in Section 6.\footnote{This research was conducted in academic year 2024/25 as part of a bachelor's thesis.}

\section{Related Work}

Spin is defined as the distorted presentation of study results, which misleads readers about how effective a treatment actually is~\cite{chiu2017spin}. It may be intentional or simply a consequence of negligence~\cite{boutron2018misrepresentation}. In RCTs, the result of interest is referred to as the outcome of the trial. Since readers often judge a study by its abstract alone, it is especially prone to spin: manual audits report spin in \textbf{47--71\% of clinical trial abstracts} that fail to confirm their hypothesis, depending on the biomedical field~\cite{khan2019level, lockyer2013spin, vera2016bias, wang2021abstracts}.

According to~\textcite{yavchitz2016new}, spin falls into three main groups:

\begin{itemize}
  \item \textbf{Misleading reporting} of results: inconsistent or incomplete reporting, such as omitting an outcome that did not confirm the hypothesis, adding a new outcome that does, or changing the order of primary and secondary outcomes.
  \item \textbf{Inappropriate interpretation} of results: incorrect conclusions drawn from even correctly reported data, such as unjustified causal language or claims not based on the actual statistical results.
  \item \textbf{Inappropriate extrapolation} of results: unjustified generalization, such as extending treatment effects to a different population.
\end{itemize}

Automatic tools for detecting spin can benefit both authors in writing, as well as readers and reviewers in critically evaluating articles. \textbf{ExaCT} (2010) uses statistical models and regular expressions to extract key trial information, such as sample size, from RCT abstracts into a table~\cite{kiritchenko2010exact}. \textbf{RobotReviewer} (2015) uses a machine learning model trained on 12,800 labeled clinical trial reports to assess their level of bias~\cite{marshall2016robotreviewer}.
\textbf{DeSpin} (2020) uses fine-tuned pre-trained language models (PLMs) to detect several types of spin in RCTs, by breaking the task down into extraction, detection, and classification sub-tasks~\cite{koroleva2020despin}. One such example is \textbf{comparing the primary and reported outcomes}, needed to detect inconsistent reporting, specifically outcome switching. This is also the task addressed in this paper, using large language models, presuming that outcomes were already extracted.

\section{Methodology}

\subsection{Dataset}

Comparing primary and reported outcomes can be framed as a directed semantic similarity task in the context of biomedicine. This is crucial for automatically checking whether the stated objectives of a clinical trial are correctly reported and met. \textcite{koroleva2019measuring} built a dataset of such outcome pairs from 3,938 RCT articles available on PubMed Central\footnote{\url{https://doi.org/10.5281/zenodo.3234827}}. One human annotator, after consulting domain experts, labeled \textbf{3,043 pairs as similar (label 1) or different (label 0)}. We then deduplicated the pairs, leaving 2,796. This dataset was imbalanced, with 21.8\% of pairs labeled as similar and 78.2\% as different, as a consequence of the collection methodology. An example pair labeled as similar is the primary outcome \textit{log HbA1c}, reported as \textit{glycaemic control}. This is a broader term, and recognizing it as a match requires domain knowledge of (blood sugar) measurements. An example of a negative pair is the primary outcome \textit{blood pressure}, reported as \textit{systolic BP}, which is a more specific term.

From the deduplicated pairs, we randomly sampled 300 for our own experiments: 100 for model tuning and 200 for validation. The remaining 2,496 pairs were used for the final testing. The tuning, validation and test sets contained 31.0\%, 21.5\% and 21.4\% similar pairs respectively, compared to 21.8\% overall. Due to unstratified sampling, the tuning set is biased toward similar pairs, relative to the whole deduplicated dataset.

\subsection{Metrics}

When comparing texts, most models produce a continuous similarity score on the interval $[0, 1]$ rather than a direct label, so a threshold is needed to turn this score into a binary decision. We used the \textbf{Youden index} to determine this threshold automatically, rather than setting it manually, which would be time-consuming and prone to bias~\cite{kim2018detecting}. The Youden index $J$ is defined at each point on the ROC curve as:

\begin{equation}
  J = \text{sensitivity} + \text{specificity} - 1 = \frac{TP}{TP + FN} + \frac{TN}{TN + FP} - 1
  \label{eq:youden_index}
\end{equation}

For each model, we selected the threshold that maximizes $J$. Unlike accuracy or $F_1$ score, $J$ treats both classes equally, making it appropriate for imbalanced datasets.

We reported classification accuracy, positive-class precision, recall, and $F_1$ score as evaluation metrics, with $F_1$ as the main one.

\subsection{Baseline Models}

\textcite{koroleva2019measuring} divided baseline text similarity models into the following groups, each representing sentences differently:

\begin{enumerate}
  \item \textbf{String measures} treat sentences as sequences of characters. We evaluated the Levenshtein distance and sequence similarity based on the Ratcliff-Obershelp distance.
  \item \textbf{Lexical measures} operate on words or word parts. We evaluated lemma similarity and stem similarity.
  \item \textbf{Vector-based measures} embed text into a vector space and compare vectors using distances such as cosine distance. We evaluated spaCy and Word2Vec embeddings.
  \item \textbf{Ontology-based measures} represent language as a hierarchy of concepts and compute distances between them. We evaluated path distance, Leacock-Chodorow similarity, and Wu-Palmer similarity.
  \item \textbf{Deep learning models} use contextual representations from large pre-trained encoders. We evaluated SciBERT and BioBERT as sentence embedding models, as well as the \textbf{all-MiniLM-L12} sentence transformer\footnote{\url{https://huggingface.co/sentence-transformers/all-MiniLM-L12-v2}}.
\end{enumerate}

\subsection{Large Language Models}
To make our approach widely accessible, we selected \textbf{open-source LLMs} available on HuggingFace and ran them locally with their transformers library. Local inference avoids API costs and latency, while keeping data private in the sensitive medical domain. We ran experiments on a single NVIDIA Quadro RTX 5000 GPU with 16 GB of VRAM, which limited the model size to up to \textbf{8B parameters}. We considered only instruction fine-tuned variants, as these follow user prompts rather than simply continuing text.

We evaluated four LLMs with different characteristics:

\begin{itemize}
  \item \textbf{OLMo-7B-Instruct}\footnote{\url{https://huggingface.co/allenai/OLMo-7B-Instruct-hf}} was chosen for its complete transparency. The architecture, training data, and procedure are publicly documented and replicable. The model was trained on 2.46T tokens, with over 2\% of scientific articles.
  \item \textbf{Mistral-7B-Instruct-v0.2}\footnote{\url{https://huggingface.co/mistralai/Mistral-7B-Instruct-v0.2}} is a general-purpose LLM with slightly better benchmark results than OLMo.
  \item \textbf{BioMistral-7B-DARE}\footnote{\url{https://huggingface.co/BioMistral/BioMistral-7B-DARE}} is a variant of Mistral further trained on PubMed Central literature, giving it stronger biomedical terminology knowledge at the cost of generality.
  \item \textbf{Llama-3-8B-Instruct}\footnote{\url{https://huggingface.co/meta-llama/Meta-Llama-3-8B-Instruct}} was reportedly trained on 15T tokens and achieves the best reasoning scores.
\end{itemize}

We first tested LLMs as binary similarity classifiers by inserting a pair of outcomes into a prompt template and extracting a \textit{Yes} or \textit{No} prediction from the generated text:

\begin{mdframed}[backgroundcolor=gray!10, linecolor=black!80,
  innerleftmargin=4pt, innerrightmargin=4pt,
  innertopmargin=4pt, innerbottommargin=4pt,
  skipabove=4pt, skipbelow=4pt]
{\small
\begin{Verbatim}
<|endoftext|><|user|>
Are the following sentences semantically similar?
First sentence: {out1}
Second sentence: {out2}
Answer:
<|assistant|>
\end{Verbatim}
}
\end{mdframed}

\begin{algorithm}[h!]
  \caption{Classification via LLM token probabilities}
  \label{alg:token_probability}
  \KwIn{Problem with a finite set of answers $a_1, a_2, \ldots, a_n$}
  Construct a prompt that expects an answer to the problem\;
  Assign each possible answer to a token $t_1, t_2, \ldots, t_n$\;
  Run the model to generate the first token\;
  Extract the output value $s_i$ for each token $t_i$\;
  Compute output probabilities $p_i = \text{softmax}(s_i)$ for each token $t_i$\;
  Find the index of the highest probability $h = \arg\max(p_1, p_2, \ldots, p_n)$\;
  \KwOut{Answer with the highest output probability $a_h$}
\end{algorithm}

This text-based approach has several limitations related to LLM answer format and extraction. We therefore adopted a \textbf{probability-based approach}, outlined in Algorithm~\ref{alg:token_probability}~\cite{yu2024breaking}. It uses probabilities of the first generated token to obtain a similarity score between 0 and 1, similar to how the MMLU benchmark~\cite{hendrycks2020measuring} evaluates multiple-choice answers. In our case, these are the tokens for \textit{Yes} and \textit{No}.

\begin{figure}[h!]
  \centering
  \includegraphics[width=\linewidth]{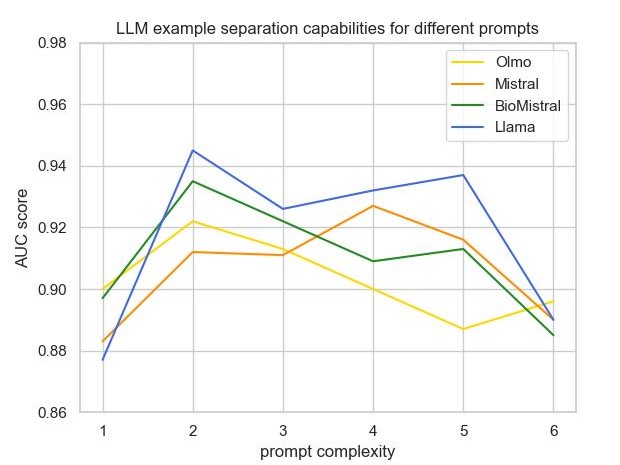}
  \caption{AUC scores across different LLMs and prompt complexity levels. Prompt names for each illustrative complexity level are provided in Table~\ref{tab:prompt_names}.}
  \Description{AUC scores across different LLMs and prompt complexity levels. Prompt names for each illustrative complexity level are provided in Table~\ref{tab:prompt_names}.}
  \label{fig:auc_vs_prompt}
\end{figure}

We used the LangChain library to define \textbf{multiple prompt templates}~\cite{chen2025unleashing}, varying in complexity from simple similarity questions to outcome switching definitions and few-shot examples. All tested prompts are available in our GitHub repository\footnote{\url{https://github.com/ta5946/Spin/blob/main/src/models/prompt_templates.py}} and can be modified as needed. Experiments used deterministic decoding with temperature set to 0. We measured AUC on the 200 instance validation set to compare model quality across prompts, before setting the classification threshold, as shown in Figure~\ref{fig:auc_vs_prompt}. 

\begin{table}[h!]
  \caption{List of evaluated prompt templates.}
  \label{tab:prompt_names}
  \centering
  \begin{tabular}{cl}
    \toprule
    Complexity level & Prompt name\\
    \midrule
    1 & Sentence template \\
    2 & Outcome template \\
    3 & Role template \\
    4 & Article definition template \\
    5 & Wikipedia definition template \\
    6 & Random example template \\
    \bottomrule
  \end{tabular}
\end{table}

The prompt with complexity 3, which defines the role of a clinical trial reviewer and the concept of primary and reported outcomes, was chosen as the starting point for further experiments. This prompt was near the middle of the AUC curve, which resembled an inverted U shape. Its average AUC score was comparable to the other prompts, but its range across models was the smallest. This indicates little bias, meaning none of them considerably over- or underperformed.

We then applied the Youden index to the 100-instance training set to determine the classification threshold and used it to obtain binary predictions on the 200-instance validation set. We \textbf{manually inspected instances misclassified} by the OLMo model and identified three recurring error types:
\begin{itemize}
  \item[(a)] insufficient domain terminology and acronym knowledge (for example, \textit{HbA1c}),
  \item[(b)] poor comprehension of longer sentences, and
  \item[(c)] acceptance of \textbf{partially reported outcomes}.
\end{itemize}
We addressed (c) by iteratively adding clarifications to the prompt, stating that the reported outcome should include all components and details of the primary outcome. Each version was re-evaluated on the validation set until it reduced this error type while preserving $F_1$ score and accuracy.

\subsection{Ensemble models}

In the final evaluation, we included ensemble models, more specifically \textbf{hard voting classifiers} with a majority threshold. Each ensemble was composed of the three models with the highest $F_1$ score on the validation set from their respective group. This improved the $F_1$ score over any single model in the group. The ensembles were:

\begin{itemize}
  \item \textbf{Baseline voting}: stem similarity, sequence similarity and sentence encoder.
  \item \textbf{LLM voting}: OLMo, BioMistral and Llama with initial prompt defining the clinical trial reviewer role.
  \item \textbf{LLM' voting}: OLMo, Mistral and BioMistral with prompt corrected to reject partially reported outcomes.
\end{itemize}

\section{Results}

\subsection{Outcome Classification}

Table~\ref{tab:model_results} shows the results of all evaluated models on the 2,496 instance test set. The sentence transformer achieved the best baseline $F_1$ score of 0.67 with a classification accuracy of 0.82. Among LLMs with the original prompt, BioMistral achieved the best $F_1$ of 0.73 and accuracy of 0.88. With the corrected prompt, OLMo' achieved the best $F_1$ of 0.75 and accuracy of 0.89. The best overall result was achieved by LLM' voting with an \textbf{$F_1$ of 0.78} and \textbf{accuracy of 0.90}, an improvement of 7 percentage points over baseline voting. \textcite{koroleva2019measuring} reported $F_1$ values between 0.87 and 0.90 with fine-tuned BERT models on a non-deduplicated dataset of outcome pairs.

\begin{table*}[t!]
  \caption{Classification thresholds and results of baseline models, LLMs, and ensemble models on the test set. LLM' denotes inference with the corrected prompt.}
  \label{tab:model_results}
  \centering
  \begin{tabular}{llccccc}
    \toprule
    Group & Model & Threshold & Classification Accuracy & Precision & Recall & $F_1$ Score\\
    \midrule
    \multirow{5}{*}{Baseline}    & Stems similarity     & 0.3      & 0.805 & 0.528 & 0.845 & 0.650 \\
                                 & Sequence similarity  & 0.4      & 0.845 & 0.668 & 0.550 & 0.603 \\
                                 & Word2Vec embeddings  & 0.4      & 0.708 & 0.409 & 0.813 & 0.544 \\
                                 & Path distance        & 0.2      & 0.795 & 0.517 & 0.673 & 0.585 \\
                                 & \textbf{Sentence transformer} & \textbf{0.4} & \textbf{0.821} & \textbf{0.553} & \textbf{0.862} & \textbf{0.673} \\
    \midrule
    \multirow{8}{*}{LLM}         & OLMo                 & 0.1      & 0.866 & 0.653 & 0.802 & 0.720 \\
                                 & Mistral              & 0.3      & 0.852 & 0.625 & 0.774 & 0.692 \\
                                 & \textbf{BioMistral}  & \textbf{0.2} & \textbf{0.881} & \textbf{0.712} & \textbf{0.751} & \textbf{0.731} \\
                                 & Llama                & 0.1      & 0.884 & 0.831 & 0.578 & 0.681 \\
                                 \cmidrule{2-7}
                                 & \textbf{OLMo$'$}     & \textbf{0.3} & \textbf{0.889} & \textbf{0.725} & \textbf{0.778} & \textbf{0.750} \\
                                 & Mistral$'$           & 0.2      & 0.894 & 0.789 & 0.692 & 0.737 \\
                                 & BioMistral$'$        & 0.1      & 0.851 & 0.609 & 0.852 & 0.710 \\
                                 & Llama$'$             & 0.1      & 0.853 & 0.962 & 0.329 & 0.490 \\
    \midrule
    \multirow{3}{*}{Ensemble}    & Baseline voting      & majority & 0.854 & 0.622 & 0.811 & 0.704 \\
                                 & LLM voting           & majority & 0.899 & 0.781 & 0.733 & 0.756 \\
                                 & \textbf{LLM$'$ voting} & \textbf{majority} & \textbf{0.901} & \textbf{0.758} & \textbf{0.793} & \textbf{0.775} \\
    \bottomrule
  \end{tabular}
\end{table*}

\subsection{Explanation}

Deep learning baseline models and the proposed LLM approach are essentially \textbf{black boxes}, producing only a similarity score with no explanation of their decisions. For authors and reviewers of clinical trial reports, understanding why two outcomes were classified as similar or different is important. Since LLMs can interpret the problem and generate text, they are suitable for producing \textbf{natural language explanations}. We obtained explanations for already classified examples by inserting the model prediction and continuing the conversation with the user message \textit{``Why? Explain your answer.''}~\cite{zhang2024language} Two examples generated by OLMo are shown below:

\begin{mdframed}[backgroundcolor=gray!10, linecolor=black!80,
  innerleftmargin=4pt, innerrightmargin=4pt,
  innertopmargin=4pt, innerbottommargin=4pt,
  skipabove=4pt, skipbelow=4pt]
{\small
\begin{Verbatim}[commandchars=\\\{\}]
Primary outcome: a reduction in depression
Reported outcome: GDS scores
Label: 1
Prediction: 1 \checkmark{}
Explanation: The reported outcome, GDS scores, are a
 measure of depression, which is the primary outcome
 in this clinical trial. The GDS is a well-known tool
 to assess depression in elderly individuals.
 Therefore, the reported outcome matches the defined
 primary outcome.

Primary outcome: pain severity and knee range of motion
 (ROM)
Reported outcome: pain
Label: 0
Prediction: 0 \checkmark{}
Explanation: The reported outcome is "pain," but the
 defined primary outcome was "pain severity and knee
 range of motion (ROM)." It does not include the
 specific components of pain severity, and it does not
 mention knee ROM. Therefore, the reported outcome
 does not match the defined primary outcome.
\end{Verbatim}
}
\end{mdframed}

\section{Discussion}

LLMs outperformed all baseline text similarity models in $F_1$, which we attribute to defining the task context and the asymmetric relationship between the primary and reported outcome. Since we used LLMs as adaptive classifiers via prompting, they remained general-purpose. BioMistral performed better with minimal prompts while Mistral benefited from more context, indicating that domain adaptation influences prompt sensitivity. Llama followed the corrected instructions too strictly~\cite{zhou2023instruction}, as its recall dropped to 0.33 and $F_1$ to 0.49, making it an unusable classifier. Individual classification thresholds, obtained by maximizing the Youden index, compensated for model biases. Majority voting across LLMs further improved results, indicating that the models made distinct mistakes and complemented each other.

Our results still \textbf{trail behind fine-tuned BERT models} by about 10 percentage points on $F_1$ score. Part of the reason is that we worked with only 10\% of the dataset for development and deduplicated instances, making the task harder. Thresholds tuned on this small, randomly sampled dataset can be unstable and vary across models. We also used LLMs to generate natural language explanations for the classified examples. A manual qualitative analysis showed that more than \textbf{50\% of explanations are valuable}, while approximately every tenth contains hallucinations.

\section{Conclusion}

We investigated whether smaller open-source LLMs can classify pairs of outcomes based on text similarity as part of detecting outcome switching in RCT reports. For this purpose, we used a dataset of primary and reported outcomes labeled for similarity. Through prompt engineering, OLMo achieved the best $F_1$ score of 0.75, and majority voting of three LLMs improved it to 0.78. Other contributions include automatic threshold tuning, classification based on next token probabilities, and the option to explain the model decision. Our approach requires only a few hundred instances and no parameter fine-tuning. All source code and results, including model predictions and explanations, are publicly available in a GitHub repository.\footnote{\url{https://github.com/ta5946/Spin}}

As future work, we propose evaluating larger, \textbf{more capable LLMs}. A complete spin detection pipeline would have to be extended to \textbf{other types of spin} and related tasks, such as outcome extraction and discrepancy judgment. Alternatively, models with longer context windows could analyze whole reports at once, replacing intermediate steps with reasoning. To ensure cross-lingual reliability, the experiments should be repeated with a machine-translated or separately collected dataset.

%% END

\printbibliography

@article{glasziou2014reducing,
  title={Reducing waste from incomplete or unusable reports of biomedical research},
  author={Glasziou, Paul and Altman, Douglas G and Bossuyt, Patrick and Boutron, Isabelle and Clarke, Mike and Julious, Steven and Michie, Susan and Moher, David and Wager, Elizabeth},
  journal={The Lancet},
  volume={383},
  number={9913},
  pages={267--276},
  year={2014},
  publisher={Elsevier}
}

@inproceedings{koroleva2020despin,
  title={{DeSpin: A} prototype system for detecting spin in biomedical publications},
  author={Koroleva, Anna and Kamath, Sanjay and Bossuyt, Patrick MM and Paroubek, Patrick},
  booktitle={Proceedings of the BioNLP 2020 workshop},
  pages={49--59},
  year={2020}
}

@article{chiu2017spin,
  title={‘Spin’in published biomedical literature: a methodological systematic review},
  author={Chiu, Kellia and Grundy, Quinn and Bero, Lisa},
  journal={PLoS Biology},
  volume={15},
  number={9},
  pages={e2002173},
  year={2017},
  publisher={Public Library of Science San Francisco, CA USA}
}

@article{yavchitz2016new,
  title={A new classification of spin in systematic reviews and meta-analyses was developed and ranked according to the severity},
  author={Yavchitz, Am{\'e}lie and Ravaud, Philippe and Altman, Douglas G and Moher, David and Hrobjartsson, Asbj{\o}rn and Lasserson, Toby and Boutron, Isabelle},
  journal={Journal of clinical epidemiology},
  volume={75},
  pages={56--65},
  year={2016},
  publisher={Elsevier}
}

@article{khan2019level,
  title={Level and prevalence of spin in published cardiovascular randomized clinical trial reports with statistically nonsignificant primary outcomes: a systematic review},
  author={Khan, Muhammad Shahzeb and Lateef, Noman and Siddiqi, Tariq Jamal and Rehman, Karim Abdur and Alnaimat, Saed and Khan, Safi U and Riaz, Haris and Murad, M Hassan and Mandrola, John and Doukky, Rami and others},
  journal={JAMA network open},
  volume={2},
  number={5},
  pages={e192622--e192622},
  year={2019},
  publisher={American Medical Association}
}

@article{vera2016bias,
  title={Bias in reporting of randomised clinical trials in oncology},
  author={Vera-Badillo, Francisco E and Napoleone, Marc and Krzyzanowska, Monika K and Alibhai, Shabbir MH and Chan, An-Wen and Ocana, Alberto and Seruga, Bostjan and Templeton, Arnoud J and Amir, Eitan and Tannock, Ian F},
  journal={European Journal of Cancer},
  volume={61},
  pages={29--35},
  year={2016},
  publisher={Elsevier}
}

@article{lockyer2013spin,
  title={“Spin” in wound care research: the reporting and interpretation of randomized controlled trials with statistically non-significant primary outcome results or unspecified primary outcomes},
  author={Lockyer, Suzanne and Hodgson, Rob and Dumville, Jo C and Cullum, Nicky},
  journal={Trials},
  volume={14},
  pages={1--10},
  year={2013},
  publisher={Springer}
}

@article{wang2021abstracts,
  title={Abstracts for reports of randomized trials of {COVID-19} interventions had low quality and high spin},
  author={Wang, Dongguang and Chen, Lingmin and Wang, Lian and Hua, Fang and Li, Juan and Li, Yuxi and Zhang, Yonggang and Fan, Hong and Li, Weimin and Clarke, Mike},
  journal={Journal of Clinical Epidemiology},
  volume={139},
  pages={107--120},
  year={2021},
  publisher={Elsevier}
}

@article{boutron2018misrepresentation,
  title={Misrepresentation and distortion of research in biomedical literature},
  author={Boutron, Isabelle and Ravaud, Philippe},
  journal={Proceedings of the National Academy of Sciences},
  volume={115},
  number={11},
  pages={2613--2619},
  year={2018},
  publisher={National Acad Sciences}
}

@article{kiritchenko2010exact,
  title={{ExaCT: Automatic} extraction of clinical trial characteristics from journal publications},
  author={Kiritchenko, Svetlana and De Bruijn, Berry and Carini, Simona and Martin, Joel and Sim, Ida},
  journal={BMC medical informatics and decision making},
  volume={10},
  pages={1--17},
  year={2010},
  publisher={Springer}
}

@article{marshall2016robotreviewer,
  title={{RobotReviewer: Evaluation} of a system for automatically assessing bias in clinical trials},
  author={Marshall, Iain J and Kuiper, Jo{\"e}l and Wallace, Byron C},
  journal={Journal of the American Medical Informatics Association},
  volume={23},
  number={1},
  pages={193--201},
  year={2016},
  publisher={Oxford University Press}
}

@article{koroleva2019measuring,
  title={Measuring semantic similarity of clinical trial outcomes using deep pre-trained language representations},
  author={Koroleva, Anna and Kamath, Sanjay and Paroubek, Patrick},
  journal={Journal of Biomedical Informatics},
  volume={100},
  pages={100058},
  year={2019},
  publisher={Elsevier}
}

@article{kim2018detecting,
  title={Detecting random responders with infrequency scales using an error-balancing threshold},
  author={Kim, Dale S and McCabe, Connor J and Yamasaki, Brianna L and Louie, Kristine A and King, Kevin M},
  journal={Behavior research methods},
  volume={50},
  pages={1960--1970},
  year={2018},
  publisher={Springer}
}

@article{naveed2025comprehensive,
  title={A comprehensive overview of large language models},
  author={Naveed, Humza and Khan, Asad Ullah and Qiu, Shi and Saqib, Muhammad and Anwar, Saeed and Usman, Muhammad and Akhtar, Naveed and Barnes, Nick and Mian, Ajmal},
  journal={ACM Transactions on Intelligent Systems and Technology},
  volume={16},
  number={5},
  pages={1--72},
  year={2025},
  publisher={ACM New York, NY}
}

@article{hendrycks2020measuring,
  title={Measuring massive multitask language understanding},
  author={Hendrycks, Dan and Burns, Collin and Basart, Steven and Zou, Andy and Mazeika, Mantas and Song, Dawn and Steinhardt, Jacob},
  journal={arXiv preprint arXiv:2009.03300},
  year={2020}
}

@inproceedings{yu2024breaking,
  title={Breaking the Ceiling of the {LLM} Community by Treating Token Generation as a Classification for Ensembling},
  author={Yu, Yao-Ching and Kuo, Chun Chih and Ziqi, Ye and Yucheng, Chang and Li, Yueh-Se},
  booktitle={Findings of the Association for Computational Linguistics: EMNLP 2024},
  pages={1826--1839},
  year={2024}
}

@article{chen2025unleashing,
  title={Unleashing the potential of prompt engineering for large language models},
  author={Chen, Banghao and Zhang, Zhaofeng and Langren{\'e}, Nicolas and Zhu, Shengxin},
  journal={Patterns},
  volume={6},
  number={6},
  year={2025},
  publisher={Elsevier}
}

@inproceedings{zhang2024language,
  title={How Language Model Hallucinations Can Snowball},
  author={Zhang, Muru and Press, Ofir and Merrill, William and Liu, Alisa and Smith, Noah A},
  booktitle={International Conference on Machine Learning},
  pages={59670--59684},
  year={2024},
  organization={PMLR}
}

@article{zhou2023instruction,
  title={Instruction-following evaluation for large language models},
  author={Zhou, Jeffrey and Lu, Tianjian and Mishra, Swaroop and Brahma, Siddhartha and Basu, Sujoy and Luan, Yi and Zhou, Denny and Hou, Le},
  journal={arXiv preprint arXiv:2311.07911},
  year={2023}
}

\end{document}